\documentclass[conference]{IEEEtran}
\IEEEoverridecommandlockouts

\usepackage{cite}
\usepackage{amsmath,amssymb,amsfonts}
\usepackage{graphicx}
\usepackage{textcomp}
\usepackage{xcolor}
\usepackage{algorithm}
\usepackage{algpseudocode}
\usepackage{float}
\usepackage{booktabs}
\usepackage{multirow}
\usepackage{qcircuit}
\usepackage{braket}           
\usepackage{url}
\usepackage{hyperref}
\usepackage{adjustbox} 

\usepackage{booktabs}
\usepackage[dvipsnames]{xcolor}

\usepackage{amsthm}

\theoremstyle{definition}
\newtheorem{definition}{Definition}
\theoremstyle{plain}
\newtheorem{proposition}{Proposition}

\def\BibTeX{{\rm B\kern-.05em{\sc i\kern-.025em b}\kern-.08em
    T\kern-.1667em\lower.7ex\hbox{E}\kern-.125emX}}
\begin{document}

\title{QIBONN: A Quantum-Inspired Bilevel Optimizer for Neural Networks on Tabular Classification}

\author{%
\IEEEauthorblockN{Pedro Chumpitaz-Flores\textsuperscript{*}}
\IEEEauthorblockA{%
\textit{University of South Florida}\\
Tampa, FL, USA\\
pedrochumpitazflores@usf.edu}
\and
\IEEEauthorblockN{My Duong\textsuperscript{*}}
\IEEEauthorblockA{%
\textit{University of South Florida}\\
Tampa, FL, USA\\
myduong@usf.edu}
\and
\IEEEauthorblockN{Ying Mao}
\IEEEauthorblockA{%
\textit{Fordham University}\\
New York, NY, USA\\
ymao41@fordham.edu}
\and
\IEEEauthorblockN{Kaixun Hua}
\IEEEauthorblockA{%
\textit{University of South Florida}\\
Tampa, FL, USA\\
khua@usf.edu}
\thanks{\textsuperscript{*}Equal  Contribution. 
}%
}

\maketitle
\begin{abstract}
Hyperparameter optimization (HPO) for neural networks on tabular data is critical to a wide range of applications, yet it remains challenging due to large, non-convex search spaces and the cost of exhaustive tuning. We introduce the Quantum-Inspired Bilevel Optimizer for Neural Networks (QIBONN), a bilevel framework that encodes feature selection, architectural hyperparameters, and regularization in a unified qubit-based representation. By combining deterministic quantum-inspired rotations with stochastic qubit mutations guided by a global attractor, QIBONN balances exploration and exploitation under a fixed evaluation budget. We conduct systematic experiments under single-qubit bit-flip noise (0.1\%--1\%) emulated by an IBM-Q backend. Results on 13 real-world datasets indicate that QIBONN is competitive with established methods, including classical tree-based methods and both classical/quantum-inspired HPO algorithms under the same tuning budget. 
\end{abstract}

\begin{IEEEkeywords}
Quantum-Inspired Algorithms, Neural Networks, Hyperparameter Optimization
\end{IEEEkeywords}

\section{Introduction}

Hyperparameter tuning is the process of systematically selecting optimal hyperparameters settings, such as learning rate, number of hidden layers, number of neurons within layers, or regularization, that control machine learning (ML) algorithms' learning processes. It is a common and crucial practice that directly impacts models' ability to generalize from training data to unseen data accurately and efficiently. Proper hyperparameter optimization (HPO) can transform algorithms into robust, high-performing models that often outperform other enhancements, which is important for tabular datasets \cite{klein2019tabular, schneider2023multi}, as common ML models, especially Neural Network (NN) variants are sensitive to different hyperparameter settings \cite{liao2022empirical}.

Aside from popular HPO algorithms like Grid Search and Random Search \cite{bergstra_random_2012}, Bayes-based optimization methods \cite{bergstra2011algorithms, victoria_automatic_2021}, evolutionary algorithms \cite{katoch_review_2021}, and population-based metaheuristic optimization \cite{kennedy_particle_1995, loshchilov2016cma}, quantum-inspired techniques for NN hyperparameter tuning have gained attention over the past few years \cite{yang_quantum_2004, lentzas2019hyperparameter, 1134125} due to their physics-based dynamics that allow for broader exploration of the search space. 
In this context, \textbf{quantum-inspired (QI) refers to classical algorithms that adopt mathematical formalism from quantum mechanics}, but do not require quantum hardware or quantum simulators; they are explicitly designed for execution on conventional digital machines \cite{zhang2011quantum, 1134125}. Recently, numerous QI algorithms such as Quantum Particle Swarm Optimization (QPSO) \cite{yang_quantum_2004}, 
Quantum-Inspired Boltzmann Machine (QIBM) \cite{ amin2018quantum}, and Quantum-Inspired Evolutionary Algorithms (QIEA) \cite{8090826} have emerged as prominent candidates for HPO, combining quantum-mechanics-inspired exploration with classical global heuristics to search more effectively in rugged, mixed discrete–continuous spaces and without requiring quantum hardware. 
Although QI approaches are theoretically applicable to NN hyperparameter tuning\cite{gan_quantum-inspired_2023, han_quantum_2023, sagingalieva_hybrid_2023}, usage for joint feature selection and neural hyperparameter search in tabular classification remains limited.
Motivated by this, 
we introduce \textbf{Quantum-Inspired Bilevel Optimizer for Neural Networks (QIBONN)}, a framework that enhances the HPO process by integrating QI concepts to increase exploration diversity and reduce the risk of premature convergence of the hyperparameter search across both shallow and deep neural architectures. We formalize the qubit update scheme (Section~\ref{sec:qi_optimization} and \ref{sec:qiht}), benchmark against classical and tabular baselines, and test the robustness of QIBONN to simulation noise (Section~\ref{sec:experiments}).

\begin{figure}[htbp]
    \centering
\includegraphics[width=0.45\textwidth]{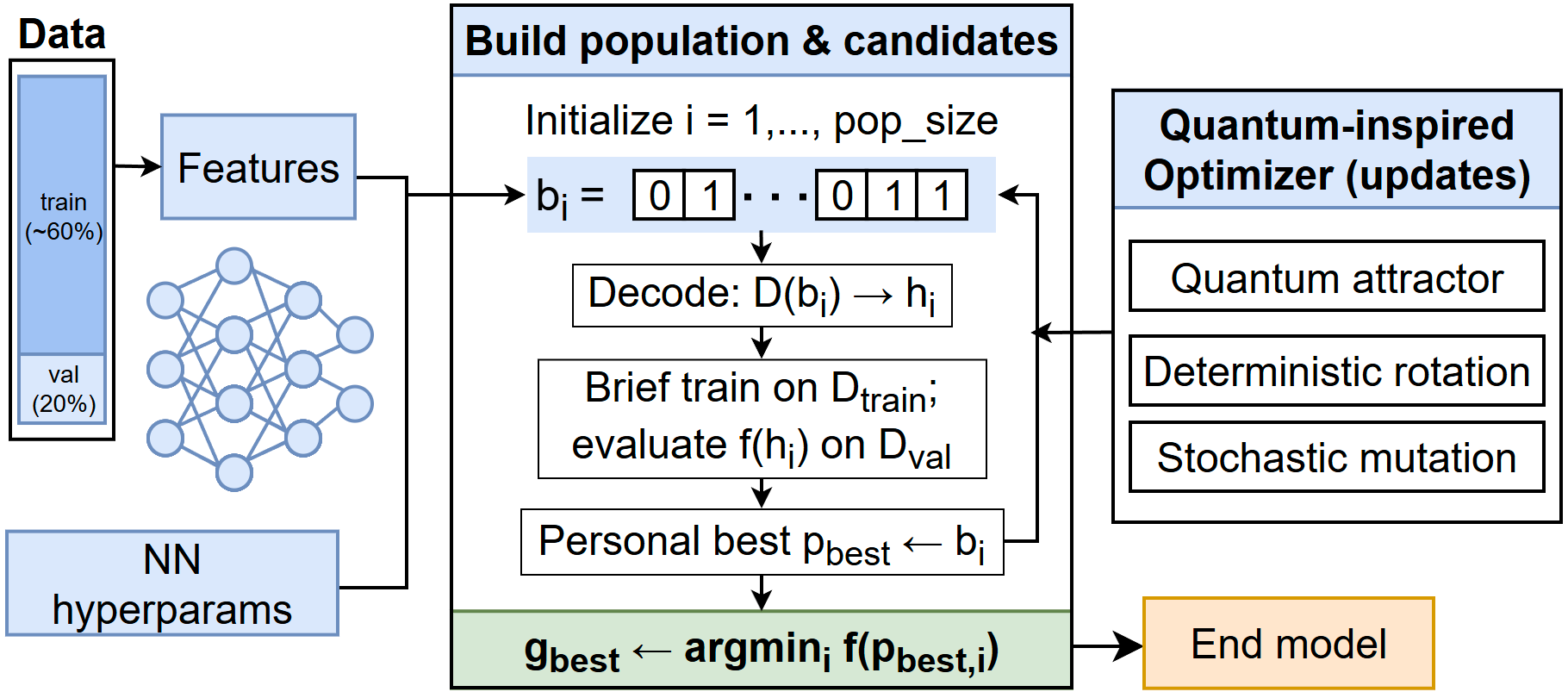} 
    \caption{QIBONN pipeline}
    \label{fig:noise_auc}
\end{figure}

\section{Quantum-Inspired Optimization}
\label{sec:qi_optimization}
The use of quantum principles has led to optimization approaches that simulate quantum dynamics at an algorithmic level without requiring real quantum hardware. Each candidate solution can be encoded as a hyperparameter vector in a qubit representation. 
Each qubit is defined as:
\[
|\psi\rangle = \alpha|0\rangle + \beta|1\rangle \quad \text{with} \quad |\alpha|^2 + |\beta|^2 = 1,
\]
where \(\alpha\) and \(\beta\) are complex amplitudes denoting the probabilities of the basis states. Superposition allows each qubit to represent multiple states simultaneously, providing a compact representation of potential solutions in high-dimensional spaces. Measurement collapses the qubit into one of the basis states with probabilities determined by the amplitudes:
\(
P(|0\rangle) = |\alpha|^2, \quad P(|1\rangle) = |\beta|^2.
\)
Unitary operators \(U\) satisfying \(U^\dagger U = I\) are used to manipulate the qubit amplitudes. A fundamental unitary operator is the rotation, defined by:
\[
R(\Delta\theta)=\begin{pmatrix}\cos(\Delta\theta)&-\sin(\Delta\theta)\\ \sin(\Delta\theta)&\cos(\Delta\theta)\end{pmatrix}.
\]
Applying \(R(\Delta\theta)\) to \(|\psi\rangle\) yields:
\[
|\psi'\rangle=\begin{pmatrix}\alpha\cos(\Delta\theta)-\beta\sin(\Delta\theta) \\ \alpha\sin(\Delta\theta)+\beta\cos(\Delta\theta)\end{pmatrix}.
\]
This rotation changes the amplitudes and, consequently, the probabilities of the basis states. The rotation angle \(\Delta\theta\) is determined by the best solutions found, thereby guiding the search toward promising regions. In addition to the deterministic rotation, quantum mutation is applied as a stochastic operator 
\(
\theta_{\text{mut}} \sim U(-\theta_{\max}, \theta_{\max}),
\)
where \(\theta_{\max}\) is the maximum allowed rotation angle. This mutation introduces diversity into the search process. 



Consider an initial qubit state \(|\psi\rangle\) that undergoes a deterministic rotation \(R_y(\Delta\theta)\), followed by a stochastic rotation \(R_y(\theta_{\text{mut}})\) applied with probability \(P_{\text{mut}}\); the qubit is then measured to yield a classical bit.
Each particle is represented by a register of qubits, with one qubit per dimension of the search space. Thus, for a problem with \(N\) particles and dimensionality \(D\), the total number of qubits is $N \times D$.



\section{Quantum-Inspired Bilevel Optimizer}
\label{sec:qiht}

Optimizing hyperparameters of neural networks for tabular data can be effectively formulated as a bilevel optimization problem. At the upper level, the goal is to identify the combination that best minimize the validation loss. At the lower level, given a set of hyperparameters, model weights are trained via standard gradient optimization. Importantly, this selection is illustrative and can include other hyperparameters like epochs, dropout rates, batch size, or any configuration relevant to the training and architecture of neural networks.

Quantum-Inspired Bilevel Optimizer for Neural Networks (QIBONN) maintains a population of qubit-encoded candidates, each endowed with its own best attractor, contributing to a shared global best attractor. These mechanisms, inherited from swarm intelligence \cite{Survey_Selvaraj_2020,Bonabeau_1990}, guide deterministic rotations and stochastic mutations of the qubits to balance exploration and exploitation. In each iteration, (i) a population of qubit-encoded candidates is generated and decoded for evaluation via a training loop; (ii) personal and global best configurations are updated; and (iii) qubit amplitudes are modified using the update operators. The details of each step are provided below.

\begin{definition}[Hyperparameter Space]
Define the hyperparameter space for \(n\) hyperparameters as:
\[
H = \left\{ \mathbf{x} \in \mathbb{R}^n \;\middle|\;
h_{i,\min} \le x_i \le h_{i,\max} \;\; \text{for all } i = 1,\dots,n \right\}.
\]
Each \(h = (x_1, x_2, \dots, x_n) \in H\) specifies a network configuration through a set of hyperparameters depending on the specific task and model architecture.
\end{definition}

\begin{definition}[Candidate Encoding and Decoding]
Let \texttt{bpp} denote the number of bits per hyperparameter. A candidate solution is represented by a bitstring:
\(
b \in \{0,1\}^{n \cdot \texttt{bpp}}.
\)
The decoding function \(D\) partitions \(b\) into \(n\) segments, converting each segment to an integer \(v_i \in \{0,\dots,2^{\texttt{bpp}}-1\}\), and maps it linearly onto the corresponding hyperparameter range:
\[
D(b)_i = h_{i,\min} + \frac{v_i}{2^{\texttt{bpp}}-1}(h_{i,\max} - h_{i,\min}), \quad \forall i = 1,\dots,n.
\]
\end{definition}

\begin{definition}[Evaluation Function]
Let \(\theta\) denote network weights. For hyperparameters \(h\in H\), let
\(
J\colon H \to \mathbb{R}, \ J(h) = -\mathcal{M}(h),
\)
where \(\mathcal{M}(h)\) is a performance metric. A network \(f(x;\theta)\) is built with architecture \(h\), weights \(\theta^*(h)\) are obtained via training, and \(\mathcal{M}(h)\) is evaluated on a validation set. The optimal hyperparameters satisfy
\(
h^* = \arg\min_{h \in H} J(h).
\)
\end{definition}

\begin{definition}[Training Phase]
After \(h^*\) is found, final weights solve
\(
\theta^* = \arg\min_{\theta} L(\theta; h^*),
\)
where \(L(\theta; h^*)\) is the train loss.
\end{definition}

\begin{proposition}[Decoupling of Hyperparameter Tuning and Weight Optimization]
For each \(h\in H\), the inner problem
\(
\theta^*(h) = \arg\min_{\theta} L(\theta; h)
\)
is solved independently, and the outer problem
\(
\min_{h \in H} J(h)
\)
depends solely on validation performance \(J(h)\).
\end{proposition}

\begin{proposition}[Quantum-Inspired Update in the Encoded Space]
Candidates \(b \in \{0,1\}^{n\cdot\texttt{bpp}}\) are updated via quantum-inspired operators altering qubit amplitudes, yielding \(b'\). Decoding maps
\(
h = D(b) \;\to\; h' = D(b').
\)
Such update drives exploration while decoupling from weight training.
\end{proposition}

At each iteration of QIBONN, we track each candidate’s personal best and the overall global best. We then compute the \emph{quantum attractor}, defined as
\[
m_{\mathrm{best}}(t) \;=\; \frac{1}{N} \sum_{i=1}^{N} x_{p\_{\mathrm{best},i}}(t),
\]
which serves as a statistical center of the search. The difference between a candidate’s current hyperparameters and the attractor finds a deterministic rotation angle for its qubit update, steering the search toward regions of high promise. To preserve exploration, we also apply stochastic rotations randomly sampled in \([-\theta_{\max},\theta_{\max}]\) with probability \(P_{\mathrm{mut}}\), preventing premature convergence.

The entire optimization workflow is outlined in Algorithm~\ref{alg:small_qiht}, which iteratively decodes qubit-encoded hyperparameters, evaluates configurations through brief model training, updates personal and global bests, and refines the search space via quantum-inspired rotations and mutations. After a fixed number of iterations, the best-found hyperparameters are decoded and used to train the final model.

\begin{algorithm}[htbp]
\caption{Quantum‐Inspired Bilevel Optimizer}
\label{alg:small_qiht}
\begin{algorithmic}[1]
\Procedure{QIBONN}{%
  $D, f, \dim, \mathrm{pop\_size}, \mathrm{max\_iter},%
   \beta, \allowbreak P_{\mathrm{mut}} ,\theta_{\max}$%
}
  \State Split dataset \(D\) into \(D_{\mathrm{train}}\) and \(D_{\mathrm{val}}\).
  \For{$i = 1$ \textbf{to} pop\_size}
    \State Generate qubit‐encoded bitstring \(b_i\).
    \State Decode \(b_i\) into hyperparameters \(h_i\).
    \State Briefly train on \(D_{\mathrm{train}}\) and evaluate \(f(b_i)\) on \(D_{\mathrm{val}}\).
    \State $p_{\mathrm{best},i} \gets b_i$.
  \EndFor
  \State $g_{\mathrm{best}} \gets \arg\min_i f(p_{\mathrm{best},i})$.
  \For{$t=1$ \textbf{to} max\_iter}
    \For{each candidate $i$}
      \State Obtain $b_i$, decode to $h_i$, train \& evaluate $f(b_i)$.
      \If{$f(b_i) < f(p_{\mathrm{best},i})$}
        \State $p_{\mathrm{best},i} \gets b_i$.
      \EndIf
      \If{$f(b_i) < f(g_{\mathrm{best}})$}
        \State $g_{\mathrm{best}} \gets b_i$.
      \EndIf
    \EndFor
    \State Compute quantum attractor $m_{\mathrm{best}}(t)$.
    \For{each qubit in every candidate}
      \State Apply $R(\Delta\theta)$, $\theta_{\text{mut}}$  with $m_{\mathrm{best}}(t)$ and $g_{\mathrm{best}}$.
    \EndFor
  \EndFor
  \State Decode $g_{\mathrm{best}}$ to obtain $h^*$.
  \State Train final model on full dataset $D$ using $h^*$.
  \State \Return final model and $h^*$.
\EndProcedure
\end{algorithmic}
\end{algorithm}

The theoretical framework is universal, supporting any total number of hyperparameters
$
n = n_{\text{feat}} + p,\quad p \ge 0,
$
where \(n_{\text{feat}}\) denotes the number of input features and \(p\) is the number of additional hyperparameters, and any bit‐precision per hyperparameter \(\texttt{bpp}\ge1\).    

\begin{table*}[htbp]
\caption{QIBONN across Shallow and Deep NN architectures vs.\ baseline vanilla NN (VNN) and GBDT on tabular datasets.}
\label{tab:qiht_boosting}
\centering
\begin{small}
\begin{adjustbox}{width=\textwidth}
\begin{sc}
\begin{tabular}{l|ccccc|ccccc}
\toprule
\multicolumn{1}{c|}{} 
  &  \multicolumn{5}{c}{ROC-AUC}&  \multicolumn{5}{c}{PR-AUC}\\
Dataset 
  &  VNN &Shallow & DeepMLP & ResMLP &Boosting 
  &  VNN&Shallow & DeepMLP & ResMLP &Boosting \\
\midrule
Cleveland&  0.7781&\textbf{0.9445}& 0.9353&       0.9317&0.8420&  0.7776&\textbf{0.9347}& 0.9016
&      0.8972&0.7870 \\
Pima Diabetes&  0.8550&\textbf{0.8926}& 0.8902
&       0.8897&0.8648&  0.7457&0.7639 
& 0.7455&      \textbf{0.8005}&0.7775 \\
German &  0.6412&0.8315& 0.8313
&       \textbf{0.8348}&0.7249&  0.4152&0.6476& 0.6804
&      \textbf{0.7572}&0.5641 \\
Telco Customer&  0.7769&\textbf{0.8653}
& 0.8521
&       0.8544&0.8535&  0.5329&\textbf{0.6905}
& 0.6591
&      0.6585&0.6822\\
     Bank Customer&  0.8034&0.8703& 0.8760&       \textbf{0.8769}&0.8671&  0.5973&\textbf{0.7353}
& 0.7151
&      0.7255&0.7102\\
Bank Marketing&  0.8317&0.8560& 0.8572&       0.8557&\textbf{0.8697}&  0.7706&0.8166& 0.8161&      0.8117&\textbf{0.8268}\\
Credit Default&   0.7231&\textbf{0.7814} & 0.7791 &      \textbf{0.7814}&0.7800&  0.4828&0.5391& 0.5443 &      0.5466&\textbf{0.5609}\\
Adult Income&   0.8908&0.9072          & 0.9046 &      0.9049&\textbf{0.9238}&  0.7497&0.7902 & 0.7783 &      0.7704&\textbf{0.8254}\\
\bottomrule
\end{tabular}
\end{sc}
\end{adjustbox}
\end{small}
\end{table*}

\section{Computational Experiments}
\label{sec:experiments}

We implemented QIBONN in Python~3.10.12 using Qiskit~1.3.2 (quantum-inspired operators) and PyTorch~2.5.1 on three architectures: a three-layer network (\emph{Shallow}), a deep MLP (\emph{DeepMLP}), and a residual MLP (\emph{ResMLP}). For qubit updates we use a quantum-inspired PSO variant~\cite{Sun_2004_pso,luitel_quantum_2010} that sets rotation angles from personal and global bests. Quantum simulations run on Qiskit’s \texttt{AerSimulator} with IBM-Q noise emulators. Experiments use a Linux server (Ubuntu kernel 6.8.0-51-generic) with an Intel\textsuperscript{\textregistered} Xeon\textsuperscript{\textregistered} Gold~6230R @ 2.10\,GHz (104 logical cores) and 187\,GiB RAM. Although the general framework in Section~\ref{sec:qiht} allows arbitrary hyperparameters and precision, our particle dimension is $(n_{\mathrm{feat}}+6)$: the first $n_{\mathrm{feat}}$ coordinates are thresholded to a binary feature mask; the remaining six map to dropout $p \in [0,0.5]$, hidden width $h \in \{8,\dots,64\}$, learning rate $\eta \in [10^{-4},10^{-1}]$ (log scale), batch size $\{32,48,64,96,128,192,256,384\}$, weight decay $\lambda \in [10^{-6},10^{-2}]$ (log scale), and hidden-layer count $L \in \{1,2,3,4\}$. Each iteration decodes a candidate, runs a short training loop to obtain validation loss, and updates qubits via deterministic rotations plus stochastic mutations around the quantum attractor. After 50 iterations, the best hyperparameters train the final model for 10 epochs on the full dataset (to keep budgets comparable).

We evaluate on eight public datasets: six from the UCI Machine Learning Repository~\cite{Dua:2019} and two Kaggle datasets (Telco Customer Churn, Bank Customer Churn). Datasets are grouped by sample count \(s\) as small (\(s \le 1{,}000\)), medium (\(s \le 10{,}000\)), and large (\(s > 10{,}000\)). HPO baselines span classical methods (grid search, random search; Bayesian surrogates via Optuna and HyperOpt), evolutionary algorithms (GP, SGA), and quantum-inspired methods (QIEA, QIBM), all within a shared NN training pipeline. QIBONN is run in three modes: (i) classical, (ii) bit-flip noise on the \textit{ry} gate with probability \(p\), and (iii) IBM emulator simulations. We report ROC-AUC and PR-AUC, averaged over 10 independent runs, and compare against an untuned vanilla NN (VNN) and a \emph{Boosting} baseline, which is the best method per-dataset among gradient boosted decision tree (GBDT) algorithms, namely XGBoost, LightGBM and CatBoost. The complete code can be found at \url{https://anonymous.4open.science/r/QIBONN-5C8B/}.

\subsection{Numerical Results}
Our work focuses on the quality of hyperparameter configurations and emphasizes reproducibility and extensibility within a unified tuning framework. QIBONN identifies configurations whose performance is competitive with standard optimization methods on tabular benchmarks by sequentially updating qubit-encoded candidates via deterministic rotations guided by personal and global best attractors, combined with stochastic mutations (Sections~\ref{sec:qi_optimization} and~\ref{sec:qiht}). 
Under a fixed evaluation budget, attains performance comparable to tree-boosting models on classification datasets with up to nearly 50{,}000 samples (Table~\ref{tab:qiht_boosting}).

Prior studies~\cite{borisov_deep_2022, shwartz2021tabular} showed that neural networks often underperform boosting algorithms on tabular benchmarks, attributing the gap to stronger inductive biases and reduced tuning complexity in tree ensembles. Tables~\ref{tab:qiht_boosting} and~\ref{tab:merged_model_comparison} show that, on small real-world datasets, QIBONN matches or exceeds the baseline quantum-inspired methods and frequently outperforms simple search, Bayesian optimization, and evolutionary algorithms, while remaining competitive with boosting on several datasets. As dataset size grows, QIBONN consistently narrows the performance gap to boosting by substantially improving upon the VNN baseline even in large-scale scenarios, demonstrating that well-tuned neural networks remain competitive at scale. On two datasets, boosting attains the best scores, whereas on every other datasets, QIBONN substantially improves all NN architectures over the VNN baseline and even outperforms boosting by over 10\% in some cases. 

For certain small datasets, other quantum-inspired and evolutionary HPO algorithms may slightly outperform QIBONN on shallow models, but we observe superior performance of QIBONN when applied on deeper or residual MLPs, especially for medium and large datasets. As dataset sizes increase, QIBONN’s scalability ensures efficient exploration of hyperparameter spaces without loss of solution quality, outperforming heuristic HPO methods and boosting, as shown in Table \ref{tab:merged_model_comparison}. In contrast, Bayesian optimization and evolutionary algorithms exhibit inconsistent ROC-AUC and PR-AUC across all medium-sized datasets. These findings indicate that QIBONN effectively scales with dataset size and maintains robust performance compared to alternative methods.
\begin{table*}[htbp]
\caption{Comparison of HPO methods on Small, Medium, and Large Real-world Tabular Datasets.}
\label{tab:merged_model_comparison}
\centering
\resizebox{\textwidth}{!}{%
\begin{sc}
\begin{tabular}{
  l|l|ccc|ccc|
  l|ccc|ccc
}
\toprule
\multicolumn{1}{c|}{} & \multicolumn{1}{c|}{}
  & \multicolumn{3}{c|}{ROC-AUC} 
  & \multicolumn{3}{c|}{PR-AUC}
  &
\multicolumn{1}{c|}{} 
  & \multicolumn{3}{c|}{ROC-AUC} 
  & \multicolumn{3}{c}{PR-AUC}
\\
Method& Dataset& FFNN & DeepMLP & ResMLP 
  & FFNN & DeepMLP & ResMLP
  &
Dataset& FFNN & DeepMLP & ResMLP 
  & FFNN & DeepMLP & ResMLP 
\\
\midrule
QIBONN 
& Cleveland
&0.9445&0.9353&\textbf{0.9317}&0.9347&0.9016&0.8972 &
Bank&\textbf{0.8703}&\textbf{0.8760}&\textbf{0.8769}&\textbf{0.7353}&\textbf{0.7151}&\textbf{0.7255}\\
Simple Search 
& 
s = 303
&0.8872&0.8882&0.8856&0.8829&0.8823&0.8778 &
Customer&0.8569&0.8498&0.8563&0.6841&0.6691&0.6773\\
Bayesian Opt. 
& 
d = 13&0.8851&0.8912&0.8908&0.8810&0.8860&0.8814 &
s=10{,}000&0.8560&0.8554&0.8571&0.6833&0.6820&0.6832\\
Evolutionary 
& &0.9306&0.9502&0.8684&0.9423&\textbf{0.9698}&0.8458 &
d=14&0.8347&0.8465&0.8463&0.6381&0.6820&0.6742\\
Quantum-Inspired & &\textbf{0.9622}&\textbf{0.9534}&0.8870&\textbf{0.9604}&0.9436&\textbf{0.9030}&
&0.8650&0.8389&0.8524&0.7014&0.6467&0.6820\\
\midrule
QIBONN 
& Pima
&0.8926&0.8902&\textbf{0.8897}&0.7639&0.7455&\textbf{0.8005}&
Bank&\textbf{0.8560}&0.8572&0.8557&0.8166&\textbf{0.8161}&0.8117\\
Simple Search 
& Diabetes
&0.8255&0.8202&0.8184&0.6799&0.6731&0.6708 &
Marketing&0.8495&0.8507&0.8503&0.8090&0.8106&0.8102\\

Bayesian Opt. 
& s=768
&0.8228&0.8130&0.8183&0.6808&0.6718&0.6726 &
s=11{,}162&0.8496&0.8491&0.8493&0.8115&0.8074&0.8100\\
Evolutionary 
& d=8
&0.8932&\textbf{0.8920}&0.8189&0.7731&\textbf{0.8297}&0.6744 &
d=16&0.8496&\textbf{0.8583}&\textbf{0.8701}&0.8042&0.8156&\textbf{0.8476}\\
Quantum-Inspired & &\textbf{0.8934}&0.8268&0.8457&\textbf{0.8238}&0.6971&0.7245 &
&0.8518&0.8408&0.8443&\textbf{0.8176}&0.8089&0.8060\\
\midrule
QIBONN 
& German
&\textbf{0.8315}&\textbf{0.8313}&\textbf{0.8348}&\textbf{0.6476}&\textbf{0.6804}&\textbf{0.7572}&
Credit&\textbf{0.7814}&\textbf{0.7791}&\textbf{0.7814}&\textbf{0.5391}&\textbf{0.5443}&\textbf{0.5466}\\
Simple Search 
& Credit
&0.6968&0.6926&0.6906&0.4898&0.4825&0.4836 &
Default&0.7529&0.7553&0.7568&0.5159&0.5060&0.5228\\

Bayesian Opt. 
& s=1{,}000
&0.6941&0.6881&0.6928&0.4818&0.4711&0.4796 &
s=30{,}000&0.7514&0.7574&0.7527&0.5141&0.5127&0.5073\\
Evolutionary 
& d=10
&0.7850&0.7368&0.7546&0.6133&0.6046&0.5738 &
d=23&0.7543&0.7634&0.7637&0.5275&0.5169&0.5393\\
Quantum-Inspired & &0.8062&0.7530&0.7834&0.6029&0.5657&0.6277 &
&0.7526&0.7620&0.7608&0.5012&0.5213&0.5227\\
\midrule
QIBONN 
& Telco
&\textbf{0.8653}&\textbf{0.8521}&\textbf{0.8544}&\textbf{0.6905}&0.6591&0.6585 &
Adult&\textbf{0.9072}&\textbf{0.9046}&\textbf{0.9049}&\textbf{0.7902}&\textbf{0.7783}&\textbf{0.7704}\\
Simple Search 
& Customer
&0.8382&0.8321&0.8317&0.6465&\textbf{0.6641}&0.6660 &
Income&0.8940&0.8946&0.8892&0.7556&0.7540&0.7428\\
Bayesian Opt. 
& s=7{,}032
&0.8382&0.8310&0.8324&0.6536&0.6590&0.6662 &
s=48{,}842&0.8920&0.8938&0.8936&0.7504&0.7525&0.7532\\
Evolutionary 
& d=21&0.8324&0.8406&0.8326&0.6341&0.6338&0.6276 &
d=14&0.8963&0.8888&0.8978&0.7555&0.7532&0.7559\\
Quantum-Inspired & &0.8391&0.8456&0.8523&0.6473&0.6433&\textbf{0.6815}&
&0.8966&0.8977&0.8956&0.7525&0.7575&0.7479\\
\bottomrule
\end{tabular}%
\end{sc}
}
\end{table*}

\subsection{Generalization to Multiclass Problems}
\begin{table*}[htbp]
\caption{QIBONN across Shallow and Deep NN architectures vs.\ baseline vanilla NN (VNN) and best-performing Gradient Boosting Decision Tree on multiclass ($K \geq 3$) tabular datasets.}
\label{tab:multiclass_model_comparison}
\centering
\begin{adjustbox}{width=\textwidth}
\begin{sc}
\begin{tabular}{lccc|ccccc|ccccc}
\toprule
& & & & \multicolumn{5}{c}{ROC-AUC}& \multicolumn{5}{|c}{PR-AUC}\\
Dataset & $s$ & $d$ & $K$ & VNN & Shallow & DeepMLP & ResMLP & Boosting& VNN & Shallow & DeepMLP & ResMLP & Boosting \\
\midrule
maternal & 1,014 & 6 & 3 & 0.591 & 0.815 & 0.825 & 0.855 & 0.920& 0.440 & 0.706 & 0.739 & 0.756 & 0.868\\
yeast    & 1,484 & 8 & 10 & 0.443 & 0.865 & 0.891 & 0.893 & 0.904& 0.128 & 0.545 & 0.573 & 0.564 & 0.616\\
hemi     & 1,955 & 7 & 3 & 0.511 & 0.812 & 0.812 & 0.864 & 0.926& 0.345 & 0.713 & 0.715 & 0.791 & 0.885\\
rds\_cnt & 10,000 & 4 & 3 & 0.500 & 0.519 & 0.501 & 0.525 & 0.499& 0.333 & 0.353 & 0.334 & 0.359 & 0.334\\
dry\_bean& 13,611 & 16 & 7 & 0.500 & 0.993 & 0.993 & 0.992 & 0.996& 0.143 & 0.971 & 0.969 & 0.969 & 0.984\\
\bottomrule
\end{tabular}
\end{sc}
\end{adjustbox}
\end{table*}
The proposed QIBONN framework can be directly extended to multiclass classification. The qubit-register representation continues to encode both feature selection and architectural hyperparameters, while the network head is adapted to a $K$-class softmax layer trained with categorical cross-entropy loss. This change affects only the output layer; \emph{no modifications are required to the bilevel quantum-inspired optimization routine}. 
We conduct experiments on multiclass real-world datasets, including four from the UCI Machine Learning Repository~\cite{Dua:2019} and the Hemicellulose dataset~\cite{wang_predicting_2022}. Evaluation is based on ROC-AUC and PR-AUC, both macro-averaged over classes using a one-vs-rest (OvR) scheme, and averaged over multiple random seeds under a fixed evaluation budget. QIBONN substantially improves on the baseline NN to approach the performance of boosting methods in 4 out of 5 datasets, and manages to surpass the best boosting algorithm on \texttt{rds\_cnt} in both metrics by a small margin of 3\%. 
The combination of qubit-based feature/hyperparameter encoding and bilevel search in QIBONN \emph{naturally generalizes to multiclass classification without altering the optimizer}.

\begin{figure*}[htbp]
    \centering
    \includegraphics[width=0.8\textwidth]{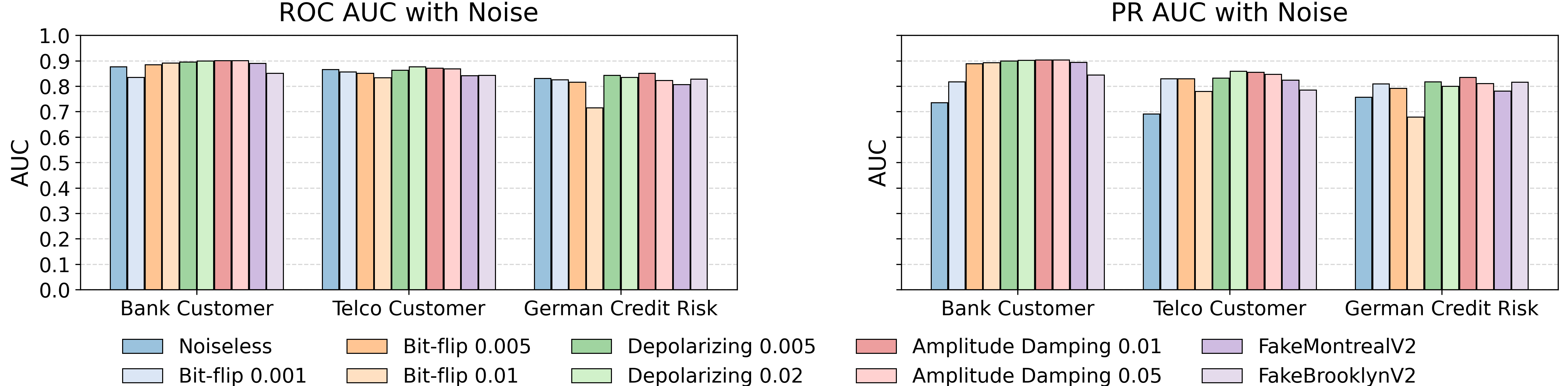} 
    \caption{Mean ROC-AUC (left) and PR-AUC (right) across three real-world tabular datasets under different noise models. Conditions are ordered as: Noiseless; Bit-flip (0.001–0.01); Depolarizing (0.005–0.02); Amplitude Damping (0.01–0.05); and IBM Q hardware emulators (FakeMontrealV2, FakeBrooklynV2).}
    \label{fig:noise_auc}
\end{figure*}


\begin{figure*}[t]
    \centering
    \includegraphics[width=0.8\textwidth]{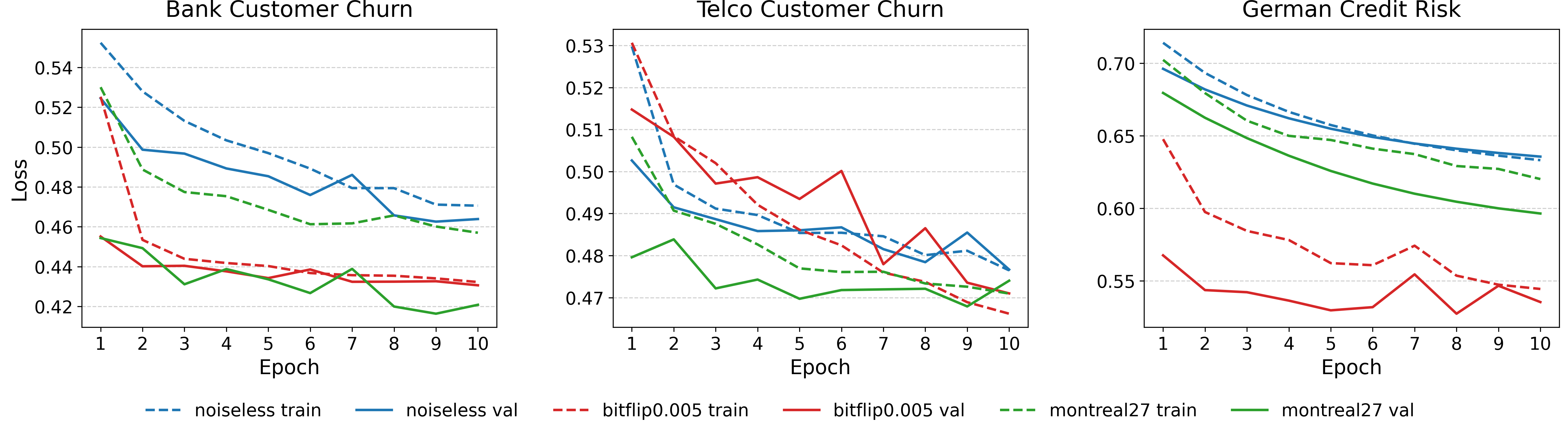} 
    \caption{Training and validation loss versus epochs for QIBONN.}
    \label{fig:training_all}
\end{figure*}

\subsection{Robustness to Noise}
To evaluate the resilience of QIBONN to qubit–level gate errors in the hardware emulators, we ran experiments on three tabular benchmarks (\emph{Bank Customer}, \emph{Telco Customer}, \emph{German Credit Risk}) under five families of conditions: (i) a noiseless \texttt{AerSimulator} baseline; (ii) single–qubit bit–flip noise with probabilities $p\in\{0.001,0.005,0.01\}$; (iii) depolarizing noise with $p\in\{0.005,0.02\}$; (iv) amplitude–damping noise with $\gamma\in\{0.01,0.05\}$; and (v) IBM Qiskit’s hardware emulators \texttt{FakeMontrealV2} (27 qubits) and \texttt{FakeBrooklynV2} (65 qubits). For the custom noise models (ii–iv), channels are injected after each state–preparation $R_y$ rotation. For the hardware emulators (v), noise is applied to the backend’s basis gates \emph{after transpilation} such as rz, sx, x, and id, to which the $R_y$ is decomposed. The fake backends reproduce gate–level calibrations (readout errors, $T_1/T_2$, gate error rates) but omit crosstalk, coherent over/under–rotations, drift, and classical–control latencies; for our single–qubit circuits the coupling map is not exercised. Figure~\ref{fig:noise_auc} reports mean ROC–AUC and PR–AUC over 5 runs; Figure \ref{fig:training_all} show loss trajectories.

Across the three datasets, Figure~\ref{fig:noise_auc} shows a mixed but consistent picture: on \emph{Bank Customer} and \emph{Telco Customer}, all noise models increase PR–AUC substantially (by +0.09–0.17 absolute) while keeping ROC–AUC typically within $\approx\!\pm 0.045$ of the noiseless baseline (max.\ $-0.042$ on \emph{Bank Customer}); the IBM hardware emulators display the same pattern on these two datasets (small ROC deltas, sizeable PR gains). On \emph{German Credit Risk}, most noise settings remain near baseline, but bit–flip $p{=}0.01$ yields a marked degradation ($\Delta$ROC $=-0.117$, $\Delta$PR $=-0.079$). Overall, QIBONN appears tolerant to \emph{moderate} noise levels (bit–flip $\le 0.005$, depolarizing $\le 0.02$, amplitude damping $\le 0.05$), with potential PR–AUC gains on two datasets, while \emph{aggressive noise} can be harmful, as seen with bit–flip $p{=}0.01$ on \emph{German Credit Risk}.
Across datasets, noisy trajectories essentially overlap the noiseless baseline in both shape and end point \emph{for moderate noise levels}, and validation curves plateau at similar values. Differences in final loss stay within the run–to–run stochastic variance, indicating no systematic degradation in convergence speed or generalization quality under these moderate conditions; by contrast, aggressive noise (such as bit–flip $p{=}0.01$ on \emph{German Credit Risk}) can degrade performance.

Taken together, these results indicate that, in simulation, \emph{moderate} single–qubit noise and hardware–emulated constraints do not induce systematic degradation, and can even improve PR–AUC on \emph{Bank Customer} and \emph{Telco Customer}; however, this should not be interpreted as evidence of universal performance gains. We view the effect as robustness to moderate perturbations rather than a reliable route to higher accuracy.

Although the noise channels are injected after each $R_y$ rotation in our custom simulations, their effect on the distribution of measured bits can be \emph{approximated} as adding a zero–mean perturbation $\xi$ at the sampling step of QPSO.
Consider the standard QPSO update \cite{Sun_2004_pso,luitel_quantum_2010}:
\[
\mathbf{x}_i(t+1)
= m_{\mathrm{best}}(t)
\;\pm\;
\alpha\,\bigl|p_{\mathrm{best},i}(t)-g(t)\bigr|\,
\ln\!\bigl(\tfrac{1}{u}\bigr),
\]
where $m_{\mathrm{best}}(t)$ is the quantum attractor, $p_{\mathrm{best},i}(t)$ the personal best of particle $i$, $g(t)$ the global best, $\alpha>0$ the step–size parameter, and $u\sim\mathcal{U}(0,1)$. 
Since $\ln(1/u)$ is exponential with parameter 
\(
\lambda = \bigl[\alpha\,\lvert p_{\mathrm{best},i}(t)-g(t)\rvert\bigr]^{-1},
\)
the displacement satisfies
\(
E[\Delta x] = \lambda^{-1},\quad 
\operatorname{Var}(\Delta x) = \lambda^{-2}.
\)
Under the measurement–noise view, we model
\[
\Delta x_{\text{noise}}=\Delta x+\xi,\qquad E[\xi]=0,\quad \operatorname{Var}(\xi)>0,
\]
with $\xi$ independent of $(u, m_{\mathrm{best}}, p_{\mathrm{best}}, g)$ at iteration $t$. Hence,
\[
E[\Delta x_{\text{noise}}]=E[\Delta x],\qquad
\operatorname{Var}(\Delta x_{\text{noise}})=\operatorname{Var}(\Delta x)+\operatorname{Var}(\xi).
\]
This variance inflation broadens the search distribution, which helps rationalize the PR–AUC gains observed on \emph{Bank Customer} and \emph{Telco Customer} at mild noise levels. We stress that this is an \emph{approximation}; it does not equate channel noise with the mutation operator, and \emph{aggressive} noise can be harmful.



\section{Conclusion}

We present QIBONN, a quantum-inspired framework for bilevel hyperparameter tuning of neural networks on tabular data. By encoding feature selection, hyperparameters, and regularization settings into a compact \((n_{\mathrm{feat}}+p)\)-qubit register, while combining the exponential QPSO update with stochastic qubit rotations, QIBONN offers a unified approach that scales linearly in the number of features. Our theoretical analysis shows that mutation-induced noise increases update variance without biasing the search, effectively acting as a light form of regularization. Empirically, on eight public datasets, QIBONN attains competitive ROC-AUC and PR-AUC relative to strong tabular baselines. Under simulation with moderate single-qubit bit-flip noise and IBM hardware emulators, we observe no systematic degradation in convergence or generalization.

\paragraph{Limitations and Future Work}\label{ssec:limitations}
Our study focuses on tabular classification tasks across a representative set of public datasets, so other tasks such as regression or scenarios with substantial missing values are not yet evaluated. The current experiments are restricted to MLPs, leaving how the method generalizes to other neural architectures open to question. Finally, results are obtained under fixed evaluation budgets and simulator-based robustness tests; broader scenarios including larger budgets, alternative stopping criteria, and runs on real quantum hardware remain directions for future work.


\color{black}
\footnotesize
\bibliographystyle{IEEEtran}
\bibliography{icml_paper}

\end{document}